\documentclass[sigconf]{acmart}

\usepackage{multirow}

\AtBeginDocument{%
  \providecommand\BibTeX{{%
    \normalfont B\kern-0.5em{\scshape i\kern-0.25em b}\kern-0.8em\TeX}}}

\begin{document}

\title{Study of Encoder-Decoder Architectures for Code-Mix Search Query Translation}
\author{Mandar Kulkarni, Soumya Chennabasavaraj, Nikesh Garera}
\affiliation{%
 \institution{Flipkart Data Science}
 \country{India}}
\email{(mandar.kulkarni, soumya.cb, nikesh.garera)@flipkart.com}

\begin{abstract}

With the broad reach of the internet and smartphones, e-commerce platforms have an increasingly diversified user base. Since native language users are not conversant in English, their preferred browsing mode is their regional language or a combination of their regional language and English. From our recent study on the query data, we noticed that many of the queries we receive are code-mix, specifically Hinglish i.e. queries with one or more Hindi words written in English (Latin) script. We propose a transformer-based approach for code-mix query translation to enable users to search with these queries. We demonstrate the effectiveness of pre-trained encoder-decoder models trained on a large corpus of the unlabeled English text for this task. Using generic domain translation models, we created a pseudo-labelled dataset for training the model on the search queries and verified the effectiveness of various data augmentation techniques. Further, to reduce the latency of the model, we use knowledge distillation and weight quantization.  Effectiveness of the proposed method has been validated through experimental evaluations and A/B testing. The model is currently live on Flipkart app and website, serving millions of queries.

\end{abstract}




\keywords{transformers, query translation}

\maketitle

\thispagestyle{fancy}
\fancyhf{}
\rhead{Accepted @ SIGIR e-Commerce Workshop 2022}

% \begin{table*}[h]
%     \centering
%     \captionsetup{justification=centering, margin=5mm}
%     \begin{tabular}{|c|c|}
%         \hline
%         \textbf{Hinglish Query} & \textbf{Issue}\\
        
%         \hline
        
%         semsung phone 8 hajar vala & Brand mispelling\\
%         \hline
        
%         laptop ke barrte & Spell errors\\
%         \hline
        
%         makhamal wala seutar & Spell errors\\
%         \hline
        
%         juta bina dori wala & Articulation Gap\\
%         \hline
        
%         teen char sau wali saree sadi & Free flowing text\\
%         \hline
        
%         kam keemat mein four g mobile & Free flowing text\\
%         \hline

%         \end{tabular}
%     \vspace{2mm}
%     \caption{Various issues associated with Hinglish queries.}
%     \label{tab:res6}
% \end{table*}

\section{Introduction}
% Customers from the non-metro cities prefer to interact with the e-commerce platform in their native language.

With the wide reach of the internet and smartphones, e-commerce platforms have an increasingly diversified user base. Due to the language barrier, native language speakers feel comfortable interacting with the platform in the language of their choice. Specifically in India, due to the relatively low percentage of the English-speaking population, the vernacular platform capability is essential to provide a reliable and convenient shopping experience to such users. 

The analysis of the in-house query data showed that, apart from English, the major portion of the queries (around 21\%) seen on the platform are code-mixed Hinglish i.e. queries where one or more Hindi words are written in Latin script. This could be because of the large Hindi-speaking population in India. To utilize traditional search workflows for English with Hinglish queries, it was important to translate Hinglish queries to English.

Code-mix translation has its unique challenges. One of the major challenges is the unavailability of a parallel training corpus. A well-known approach for improving the translation model with large target-side monolingual data is back-translation \cite{sennrich2016improving}. However, such an approach is not feasible with code-mix data since there is no publicly available model which can reliably translate from English to Hinglish. On top of it, for search query translation, factors such as spell errors in the input queries, ambiguity in language identification, non-standard query format, and free-flowing text pose additional challenges. For some queries, we also encounter articulation gap issues i.e. the user is not able to correctly articulate the query to search the product e.g. user inputting a query 'juta bina dori wala' where the intended product is 'shoe without lace'. In such cases, literal translation may not provide good search results. Table \ref{tab:res6} shows examples and issues associated with Hinglish queries.

\begin{table}[!htbp]
    \centering
    \captionsetup{justification=centering, margin=5mm}
    \begin{tabular}{|c|c|}
        \hline
        \textbf{Hinglish Query} & \textbf{Issue}\\
        
        \hline
        
        semsung phone 8 hajar vala & Brand mispelling \\
        \hline
        
        redami 5a ki battery & Brand mispelling \\
        \hline
        
        laptop ke barrte & Spell errors \\
        \hline
        
        makhamal wala seutar & Spell errors \\
        \hline
        
        juta bina dori wala & Articulation Gap \\
        \hline
        
        baal hawa machine & Articulation Gap \\
        \hline
        
        teen char sau wali saree sadi & Free flowing text \\
        \hline
        
        kam keemat mein four g mobile & Free flowing text \\
        \hline

        \end{tabular}
    \vspace{2mm}
    \caption{Various issues associated with Hinglish queries.}
    \label{tab:res6}
\end{table}

% \begin{figure*} 
% \centering
% \begin{tabular}{c}

% \includegraphics[width=400pt, height = 80pt]{encoder_decoder.png}

% \end{tabular}
% \caption{\label{fig:inpt} Transformer model for Hinglish to English translation.}
% \end{figure*}

In this paper, we propose a transformer-based approach for code-mix query translation. We propose to use the pre-trained encoder-decoder models such as T5, and BART as the base model and further fine-tune it on the in-domain query corpus. As opposed to other works on code-mix data such as \cite{jawahar2021exploring}, which proposes to use multi-lingual pre-trained models, we experiment with models trained only on English text.
% A parallel corpus for the training is created using the monolingual Hindi query data using publicly available translation and transliteration apis. 
A parallel corpus for the training is created with monolingual Hindi and Hinglish queries. To detect Hinglish queries, we developed a ML-based approach. We use  translation and transliteration APIs for quick and in-expensive labeling with unlabeled queries. We observe that  translations on the search queries are noisy, possibly due to domain mismatch. To make the translation robust against noise and issues such as spell errors, we experiment with different data augmentation schemes, and the model is trained using a combination of supervised and data augmentation objectives. To mitigate issues with the articulation gap, we use a small set of manually tagged Hinglish queries.
To reduce the latency of the model in production, we use Knowledge Distillation (KD) and Weight Quantization (WQ). Experimental results under various settings demonstrate that our approach can provide reliable translations for the queries. Post successful results with AB testing, the model is currently live in production, serving millions of queries. The main contributions of the paper are as follows.
%\vspace{-0.5cm}
\begin{itemize}
  \item Validating the effectiveness of English pre-trained models and data augmentations for code-mix search query translation which achieves 25 BLEU points improvement over generic in-house models and third-party APIs  
  \item Developing a Knowledge Distillation (KD) approach for faster inference which achieves more than 4x reduction in latency with a small drop in the accuracy
  \item Developing Hinglish language detection approach to identify code-mix queries
  
%   \item Creating a 5M parallel training corpus for code-mix query translation using  APIs
%   \item Good business impact as estimated by metric improvements observed through A/B testing: model is expected to lead to \textasciitilde 616Cr of GMV gain
%   \item Externalization: Good translation QC accuracy (96\%) and \textasciitilde 5\% search pain reduction (with SQA) observed for Myntra Hinglish queries

\end{itemize}

\section{Related work}
\pagestyle{plain}

Transformers \cite{vaswani2017attention} are widely used for seq2seq applications such as translation. Warm starting the transformer training with pre-trained checkpoints proved to be an effective method for improving the translation model. Rothe et.al \cite{rothe2020leveraging} extensively studied the effectiveness of such a method for a wide variety of Natural Language Processing (NLP) tasks. It was observed that, for encoder-decoder models, good results are observed when the encoder and the decoder share the vocabulary. 

In a recent work on code-mix translation, Jawahar et al. \cite{jawahar2021exploring} explored the use of multilingual pre-trained encoder-decoder models such as mBART and mT5 for translation from English to Hinglish. On the contrary, in our work, we explore the use of pre-trained models trained only on English text.

Due to the lack of parallel corpus for training, few of the earlier works attempted to use existing MT systems for code-mix translation. Dhar et al. \cite{dhar-etal-2018-enabling} explored augmenting code-mix data for use with existing MT systems such as Google Translate. Srivastava et al. \cite{srivastava-singh-2020-phinc} also explored the use of existing MT systems such as Google Translate and Bing for translating code-mix text. They also propose a partial translation approach based on token language labels along with MT systems.

\section{Exploring pre-trained checkpoints for query translation}

\begin{table*}[h]
    \centering
    \captionsetup{justification=centering, margin=5mm}
    \begin{tabular}{|c|c|c|c|}
        
        \hline
        \textbf{Setting} & \textbf{BART-base} & \textbf{T5-base} & \textbf{Bert2Bert}\\
        \hline
        Baseline & 31.88 & 32.8 & 19.88\\
        \hline
        +AutoEncoder & +1.37 & +1.13 & +7.13\\
        \hline
        +DropChar  & +1.32 & +1.14 & +1.8\\
        \hline
        +Masking & +1.14 & +1.43 & +5.02\\
        \hline
        +WordOrderPermute & -3.1 & -7.4 & +5.6\\
        \hline
        \end{tabular}
    \vspace{2mm}
    \caption{Comparison result for models and data augmentations. Baseline result indicates BLEU scores with pre-trained models without using data augmentation. Rows 2-5 indicate a change in the Baseline BLEU score when a particular data augmentation is used during the training.}
    \label{tab:res}
\end{table*}

Pre-trained transformer models are proved to be effective in a wide variety of Natural Language Processing (NLP) applications. 
For translating code-mix Hinglish queries to English, we experiment with different pre-trained encoder-decoder models and further fine-tune these models on the in-domain query data. The model is fine-tuned with the combination of two loss components as shown in Eq. \ref{eq:so1}. $Loss_s$ corresponds to the cross-entropy loss for the labeled set of queries while the $Loss_d$ corresponds to the cross-entropy loss for the data augmentation. 
\begin{eqnarray}\label{eq:so1}
L =  (1 - \lambda) \hspace{0.1cm} Loss_s + \lambda \hspace{0.1cm} Loss_{d}
\end{eqnarray}
where $\lambda$ indicates the weighting factor for the supervised and the data augmentation loss.

In the following sections, we describe the details of the training data, models, and data augmentations techniques experimented with for the training.

\subsection{Data preparation}
\label{sec:d}

As noted earlier, one of the major challenges with code-mix translation is the unavailability of parallel corpus for training. To identify the pre-trained model best suited for the code-mix translation task, for the initial set of experiments, we created a parallel training set of 200k queries. We used monolingual Hindi query data and transliterated it to English (to get Hinglish) and translated it to English (to get the English target label). For translation and transliteration of the monoloingual data, we use generic domain third-party APIs and also observe comparable results of the same with generic in-house models in \ref{sec:inhouse}.

\subsection{Test data preparation}

For evaluation, we created a parallel test set of 15k manually tagged Hinglish queries. The test set is formed based on the query impressions (repetitions) i.e. the test set consists of approximately equal proportions of Head, Torso, and Tail queries. The same test dataset is used for all the experimental evaluations in the paper.

% In the 200k query dataset, the median length of the queries is 4 words while 90\% of the queries have less than 6 words. 

% from the set of English queries, we identified 25k hinglish queries based on the keyword list. If any of the word in the keyword list is present in the query, query is marked as Hinglish. We created a dataset of 25k hinglish queries and got it manually tagged for translation to english. This dataset acts as test set for our application. We also developed an automated hinglish query detection approach. 

\subsection{Pre-trained models}

For query translation use-case, we explored three pre-trained encoder-decoder models.  

\subsubsection{\textbf{T5}}

Text-To-Text Transfer Transformer (T5) \cite{raffel2020exploring} is the transformer encoder-decoder model trained in a self-supervised manner on a large corpus of unlabeled English text. We use the T5-base model which has ~222M trainable parameters and 12 layers in the encoder and decoder. For our application, since T5 is fine-tuned on a single task, we do not add any pre-text to the input queries during training/inference. The model is fine-tuned for 3 epochs using AdamW optimizer with a learning rate of 1e-5 and batch size of 32.

\subsubsection{\textbf{BART}}

BART \cite{lewis2019bart} is a transformer model trained using de-noising autoencoder objective. We use the Bart-base model which has ~139M trainable parameters and 6 layers in the encoder and decoder. The model is fine-tuned for 3 epochs using AdamW optimizer with a learning rate of 5e-6 and batch size of 64.

\subsubsection{\textbf{Bert2Bert}}

Rothe et al. \cite{rothe2020leveraging} demonstrated the effectiveness of pre-trained checkpoints for sequence generation. 
We use pre-trained bert-base-uncased \cite{devlin2018bert} model as the encoder and the decoder since it showed good performance for translation task \cite{rothe2020leveraging}. Since the BERT is inherently an encoder model while using it as the decoder, additional cross attention weights are introduced and these are initialized randomly. Therefore, the encoder and the decoder both have 12 layers. The model is fine-tuned for 3 epochs using AdamW optimizer with a learning rate of 5e-6 and batch size of 32.

For all the models, we use label smoothing during the training since it is shown to provide robustness to the label noise \cite{vaswani2017attention}. The label smoothing parameter is set to 0.1 for all the experiments. During the inference, we use beam search decoding where the beam size is set to 3. All the experiments were carried out on A100 GPU. 

\subsection{Data Augmentation}

We have experimented with the following data augmentation techniques.

\subsubsection{\textbf{Masking}}
A word/token masking has proved be an effective self-supervised training strategy \cite{lewis2019bart} \cite{raffel2020exploring}. Since search queries are short in length, we randomly mask a single word from the target (English) and train the model to reconstruct the entire target. For the BART and BERT2BERT models, a randomly selected word is replaced with the '[MASK]' token while for T5, it is replaced with \texttt{'<extra\_id\_0>'} token. 

\subsubsection{\textbf{AutoEncoder}}
\label{sec:ae}

We experiment with Autoencoder (AE) as the self-supervised data augmentation objective. For this, we use target data as the input and train the model to reconstruct it. The intuition behind it is that we want to teach the model to reproduce English words in the query as it is. AE techniques have been experimented with NMT \cite{cheng2016semisupervised}, however, authors have used an additional target-to-source model to auto-encode the source sentence. In our work, we only use a single source-to-target model for the AE objective.

\subsubsection{\textbf{DropChar}}

Since non-English speakers attempt to spell the words based on the phoneme sound of it, we observe that typically the first and last character of the word is spelled correctly while the majority of the spelling mistakes are present in the middle of the word. To render translation robust to such spell mistakes, we randomly drop a character from 30-50\% of the words in the query and train the model with the corresponding translation as the target.

\subsubsection{\textbf{WordOrderPermute}}

As shown in \cite{lewis2019bart}, permuting the order of words in the sentence has proved to be an effective noise technique. In our case, we randomly permute the order of the words in the target and train the model to de-noise it. 

For all the data augmentations, for each batch of supervised labeled data, we randomly sample a batch of queries from the dataset and apply a data augmentation to it. The model is then updated with a linear combination of supervised and data augmentation loss.
For all the experiments, we set $\lambda$ to 0.5 (Eq. \ref{eq:so1}).

\subsection{Results}

To judge the effectiveness of different data augmentation techniques, we performed an experiment where we first train a Baseline model without data augmentation. Then, we add each data augmentation, one at a time, and observe its effect. This experiment is carried out for all the models. 
Table \ref{tab:res} shows the BLEU score result of the experiment. All the BLEU scores are computed using SacreBLEU \cite{post2018clarity}. 

From the results, we draw a few observations. AutoEncoder, CharDrop, and Masking provide an improvement over the baseline for all the models. Surprisingly, a simple data augmentation such as Auto-encoder provides improvement to the accuracy. We further analyze the effect of AE augmentation with additional experiments and details are provided in Appendix section \ref{sec:analy}.
WordOrderPermute has a negative effect over the baseline in the case of BART and T5. The reason could be the lack of grammar in the search queries and hence, the model is not able to identify the relative positions of the words. BART and T5 provide comparable accuracies while Bert2Bert gave inferior results. This indicates that the selection of the pre-trained model has an effect on the accuracy and some of the pre-trained embeddings are more favorable than others. 

\section{Hinglish query detection}
\label{sec:langdetect}

To detect Hinglish code-mix queries, we developed a ML based approach. We chose 80k search queries randomly and got them manually tagged for each word in the sequence. Each word is tagged as English, Hinglish or Other languages. From this, we set aside 20k search queries as a test set.

For query language detection, we explored text classification and sequence labeling approaches. For text classification, the entire query is tagged as English if all the words are English, it is tagged as Hinglish if any of the words is Hinglish and it is tagged as Other if all the words are neither English nor Hinglish. 
We tried following classification models. 1. fasttext classifier, where the word embeddings are averaged and passed through a softmax classifier. 2. 1-D CNN model with softmax classifier where filter size is set to three. 3. Single layer LSTM model with softmax classifier.

% we tag the entire query into English if all words are English, Hinglish even if one word is Hinglish, Other if the query has other language words. 

\begin{table}[h]
    \centering
    \captionsetup{justification=centering, margin=5mm}
    \begin{tabular}{|c|c|c|c|}
        
        \hline
        \textbf{Model} & \textbf{Precision} & \textbf{Recall} & \textbf{F1 score}\\
        \hline
        Fasttext & 0.90 & 0.61 & 0.73\\
        \hline
        CNN & 0.80 & 0.65 & 0.72\\
        \hline
        LSTM  & 0.82 & 0.63 & 0.71\\
        \hline
        CRF & 0.84 & 0.72 & 0.76\\
        \hline
        % CRF + Hinglish word list & 0.82 & 0.81 & 0.81\\
        % \hline
        % CRF + Dict & 0.82 & 0.81 & 0.81\\
        % \hline
        \end{tabular}
    \vspace{2mm}
    \caption{Comparison of different approaches for Hinglish query detection}
    \label{tab:res15}
\end{table}

Apart from this, we also tried the sequence labeling approach - where we utilize word-level labels in the query. We tried Conditional Random Field (CRF) with a context window of 3 - current, previous, and next word. For CRF, the features used were character n-grams (uni, bi, tri, quad), length of the word, whether the word contains digits, and whether the word contains special characters. While inference, a query is termed as Hinglish if any of the words is classified as Hinglish.

Table \ref{tab:res15} shows the comparison results for the different approaches. CRF model provides the best results compared to other approaches. 
On the test set, a simple keyword-based approach provides precision and recall of 77\% and 28\% respectively while the CRF-based approach provides recall of 72\%, thus improving the detection recall by more than 2x.

\section{Training corpus preparation}

For capturing more word variations, we prepared a training corpus derived from the incoming traffic of unlabeled Hindi and Hinglish queries. We used generic domain third-party APIs for translation and transliteration and also experimented with training in-house generic translation and transliteration models for data tagging, and the results are given in section \ref{sec:inhouse}. The parallel query training dataset is created as follows:

\begin{itemize}
\item Unlabeled Hindi queries: Translating and transliterating them to English
\item Unlabeled Hinglish queries: First transliterating to Hindi followed by Translation to English
\end{itemize}

% \begin{itemize}
% \item Unlabeled Hindi queries: Translating and transliterating them to English
% \item Unlabeled Hinglish queries: Translating it to English
% \end{itemize}

% To identify Hinglish queries, we use a simple keyword based approach.
% We manually created a set of 900+ Hinglish keywords and a query is identified as Hinglish if it contains any of these words.

Hindi (Devanagari) queries are filtered based on the unicode ranges for the script. We also observe a very small percentage of Marathi queries since it shares a script with Hindi. We did not specifically filter out Marathi queries. 
For Hinglish language detection, we use the approach mentioned in section \ref{sec:langdetect}.

We created a parallel corpus of 5M queries for training the model. We observe that translations with  are noisy possibly due to domain mismatch. Therefore, for testing and fine-tuning with cleaner labels, we also created a dataset of 70k Hinglish queries tagged in-house. Note that the in-house tagged data is less than 1.5\% of the total labeled set.

For model training, we follow a two-stage process. In the first stage, we use 5M noisy query annotations for fine-tuning the model where we warm start the training with the existing pre-trained checkpoints. We use the loss function described in Eq. \ref{eq:so1} for training the model. We use DropChar, AutoEncoder, and Word Masking as the data augmentation techniques. For each batch of supervised data, a data augmentation method is chosen randomly and applied to the randomly sampled query batch to compute $Loss_{d}$.  The model is trained for 5 epochs with the learning rate 5e-6 and batch size of 64. In the second stage, we further fine-tune the model on the in-house labeled set of 55k queries where 10\% of queries are used for validation and the remaining 15k queries are used as the test set. During the second stage of fine-tuning, we only use DropChar, and AutoEncoder data augmentations and the model is fine-tuned until the validation loss seizes to improve.  We use label smoothing for both stages where the label smoothing parameter is set to 0.1. During inference, we use beam search decoding with a beam size of 3.

\begin{table}[h]
    \centering
    \captionsetup{justification=centering, margin=5mm}
    \begin{tabular}{|c|c|}
        
        \hline
        \textbf{Model} & \textbf{BLEU}\\
        \hline
        Azure Translate & 25.5 \\
        \hline
        Google Translate & 35.8 \\
        \hline
        RandomInit BART & 49.9 \\
        \hline
        BART w/o DataAug & 50.3 \\
        \hline
        BART with DataAug & \textbf{51.3} \\
        \hline
        
        \end{tabular}
    \vspace{2mm}
    \caption{Results on Hinglish test set}
    \label{tab:res1}
\end{table}

\begin{table*}[hbt!]
    \centering
    \captionsetup{justification=centering, margin=5mm}
    \begin{tabular}{|c|c|c|c|c|}
        
        \hline
        \textbf{Model} & \textbf{\# layers} & \textbf{KD loss} & \textbf{BLEU} & \textbf{p95}\\
        \hline
        BART teacher & 6 & - & 51.3 & $\sim$ 200 ms\\
        \hline
        Student (Q.) & 1 & CE & 49.7 & $\sim$ 47 ms \\
        \hline
        Student (Q.) & 1 & JS & 50.8 & $\sim$ 47 ms\\
        \hline
        % OpenNMT Student (Q.) & 1 & - & 51.1 & $\sim$ 14 ms \\
        % \hline
        % Only OpenNMT (Q.) & 1 & - & 38.5 & $\sim$ 14 ms \\
        % \hline
        
        \end{tabular}
    \vspace{2mm}
    \caption{Results for Knowledge Distillation. (Q.) indicates the Weight Quantization in the student.}
    \label{tab:res2}
\end{table*}

In an earlier experiment, since BART and T5 gave comparable results, we experimented with both the models and computed the BLEU scores on the test set. BART-base and T5-base gave BLEU scores of 51.3 and 51.33 respectively. Interestingly, though the BART-base model is less deep, it provides comparable results T5-base which has twice the number of layers. This indicates that for the query translation, deeper models may not necessarily provide an advantage possibly due to the lack of grammar in the queries. We choose BART for further experiments because it gave comparable results to T5.  

Further, to validate the effectiveness of warm starting the training with pre-trained weights, we did an additional experiment. We randomly initialize the BART model with weights as $\mathcal{N}(0,0.02)$ and follow the two-stage training procedure as explained earlier. Table \ref{tab:res1} shows the comparison result. The result indicates that starting the training with pre-trained weights provides better BLEU scores. Hence, pre-trained weights as well data augmentations are essential for good performance.

% Due to domain mismatch, the generic domain third-party translations are observed to be noisy. We validated this by getting translations on the Hinglish test set. 

We also compared our result with general domain translations from Azure and Google translate. Hinglish queries are first transliterated to Hindi, and then a Hindi-to-English translation is used. The first and second row in Table \ref{tab:res1} indicates the BLEU score result. The generic domain translators have significantly lower accuracies than the proposed model, possibly due to domain mismatch.

% Even though BLEU scores with translations are lower, with appropriate noise handling during training, these noisy annotations can be used for getting high-quality translations.

\begin{table*}[h]
    \centering
    \captionsetup{justification=centering, margin=5mm}
    \begin{tabular}{|c|c|c|}
        \hline
        \textbf{Hinglish Query} & \textbf{English Translation} & \textbf{Issue}\\
        
        \hline
        semsung phone 8 hajar vala & samsung phone for 8 thousand rupees & Brand mispelling\\
        \hline
        
        redami 5a ki baitiry & redmi 5a battery & Brand mispelling\\
        \hline
        
        % tel mi 79 back cover & realme 79 back cover & Brand mispelling\\
        % \hline
        
        goldan kalar ka aie linar & golden colour eye liner & Spell errors\\
        \hline
        
        lenovo k6 ka kabar & lenovo k6 cover & Spell errors\\
        \hline
        
        makhamal wala seutar & velvet sweater & Spell errors\\
        \hline
        
        resmi 7a kwr & redmi 7a cover & Spell errors\\
        \hline
        
        250 tak ki biluthuth & bluetooth upto 250 & Spell errors\\
        \hline
        
        % baal seedha karne ka & hair straightener & Articulation Gap\\
        % \hline
        
        baal hawa machine & hair dryer machine & Articulation Gap\\
        \hline
        
        juta bina dori wala & shoe without lace & Articulation Gap\\
        \hline
        
        teen char sau wali saree sadi & saree for 300-400 rupees & Free flowing text\\
        \hline
        kam keemat mein four g mobile & 4g mobile in low price & Free flowing text\\
        \hline
        
        % 12 sal ke bacche ke liye kapde bataiye & show clothes for 12 years old kids & Free flowing text\\
        % \hline
        
        % 2 saal ke bachche ka gaadi kam damon mein & car for 2 years old kids at low price & Free flowing text\\
        % \hline
        
        \end{tabular}
    \vspace{2mm}
    \caption{Example results for Hinglish to English query translations.}
    \label{tab:res5}
\end{table*}

\section{Knowledge Distillation (KD)}

For a good user experience, a quick turnaround with search queries is essential. Hence, it is important to have a low latency translation model. Due to a large number of parameters, self-attention computation across all layers, and auto-regressive generation, translation models typically have a higher latency during inference.  
To reduce the latency of the model, we use Knowledge Distillation (KD) where we train a smaller student model with the larger BART model as the teacher. In particular, we follow an approach proposed in  \cite{kim2016sequencelevel} for training the student model. We use a single-layer BART model as the student.  
We use the following loss function to train the student model.
\begin{eqnarray}\label{eq:so2}
L_{student} =  (1 - \lambda) \hspace{0.1cm} (Loss_s + Loss_d) + \lambda \hspace{0.1cm} Loss_{KD}
\end{eqnarray}
The training loss consists of three loss components: supervised cross-entropy loss, data augmentation cross-entropy loss, and Knowledge Distillation (KD) loss. $\lambda$ is set to 0.5 for all experiments. For the KD loss component, we experimented with Cross-Entropy (CE) and Jensen-Shannon (JS) loss. In contrast to Kullback–Leibler (KL) divergence or Cross-Entropy (CE), JS loss is symmetric, bounded, does not require absolute continuity \cite{englesson2021generalized}. 
The JS loss is defined as follows.
\begin{eqnarray}\label{eq:so3}
JS =  D_{KL}(T_p || m) + D_{KL}(S_p || m)
\end{eqnarray}
Where $T_p$ and $S_p$ indicate output probability distributions of the teacher and the student respectively. The distribution $m$ is obtained as follows. 
\begin{eqnarray}\label{eq:so4}
m =  0.5 * T_p + 0.5 * S_p
\end{eqnarray}
While training the student model, $Loss_s$ and $Loss_d$ are computed on the batch of manually labeled data. For each batch of manually tagged data, we randomly sample a batch of Hinglish queries from the 5M corpus. We obtain the pseudo translation labels from the teacher on this Hinglish query batch. Using input Hinglish queries and corresponding pseudo labels and with teacher forcing strategy, we obtain probability distributions for the teacher ($T_p$) and the student ($S_p$). Note that the teacher forcing term here is generic to transformers and not related to the teacher model in KD. $Loss_{KD}$ is then computed with either CE or JS. We use Weight Quantization to further speed up the inference on the CPU devices. 

\begin{figure*} [!t]
\centering
\begin{tabular}{c c}

\includegraphics[width=200pt, height = 150pt]{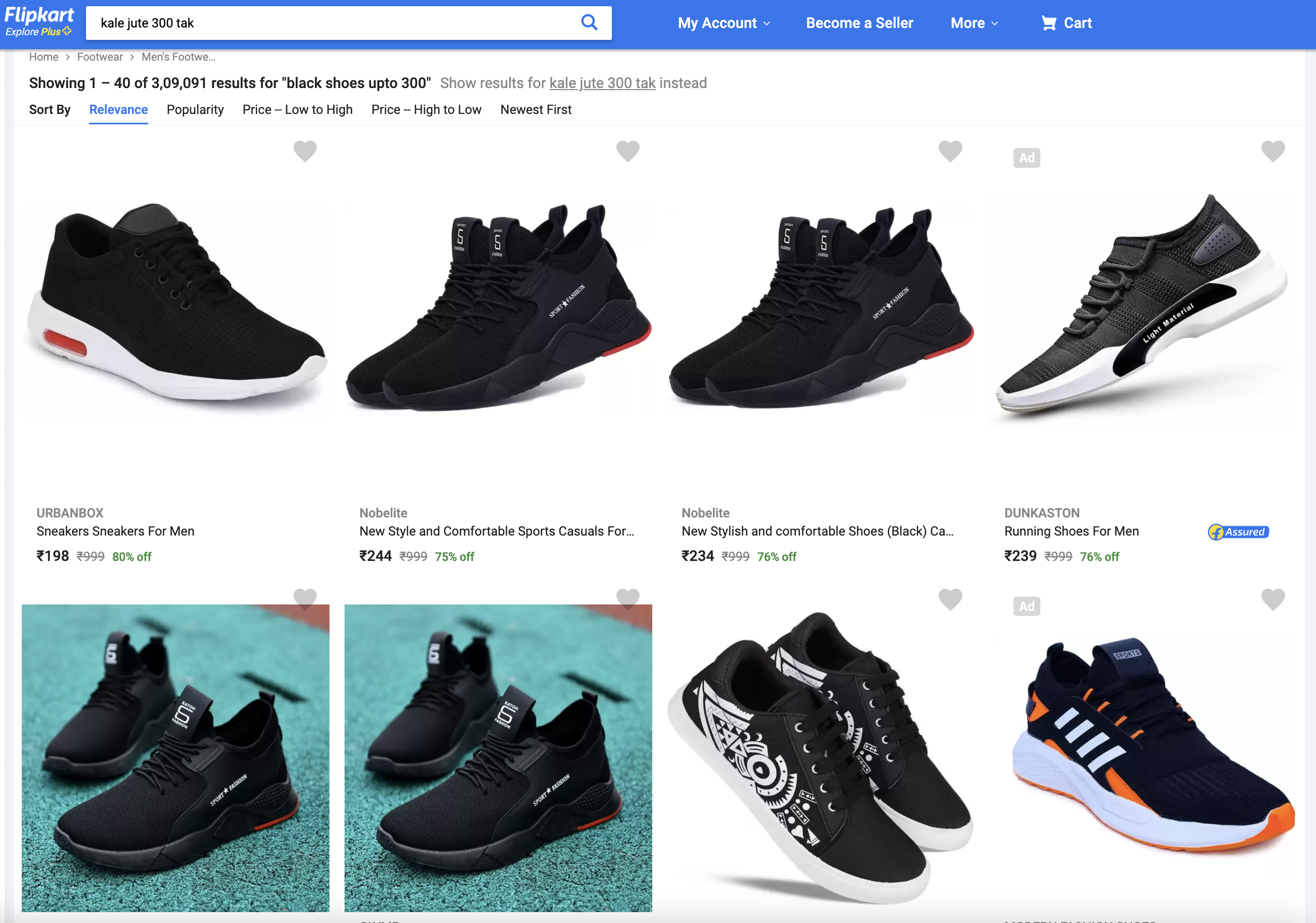}&
\includegraphics[width=200pt, height = 150pt]{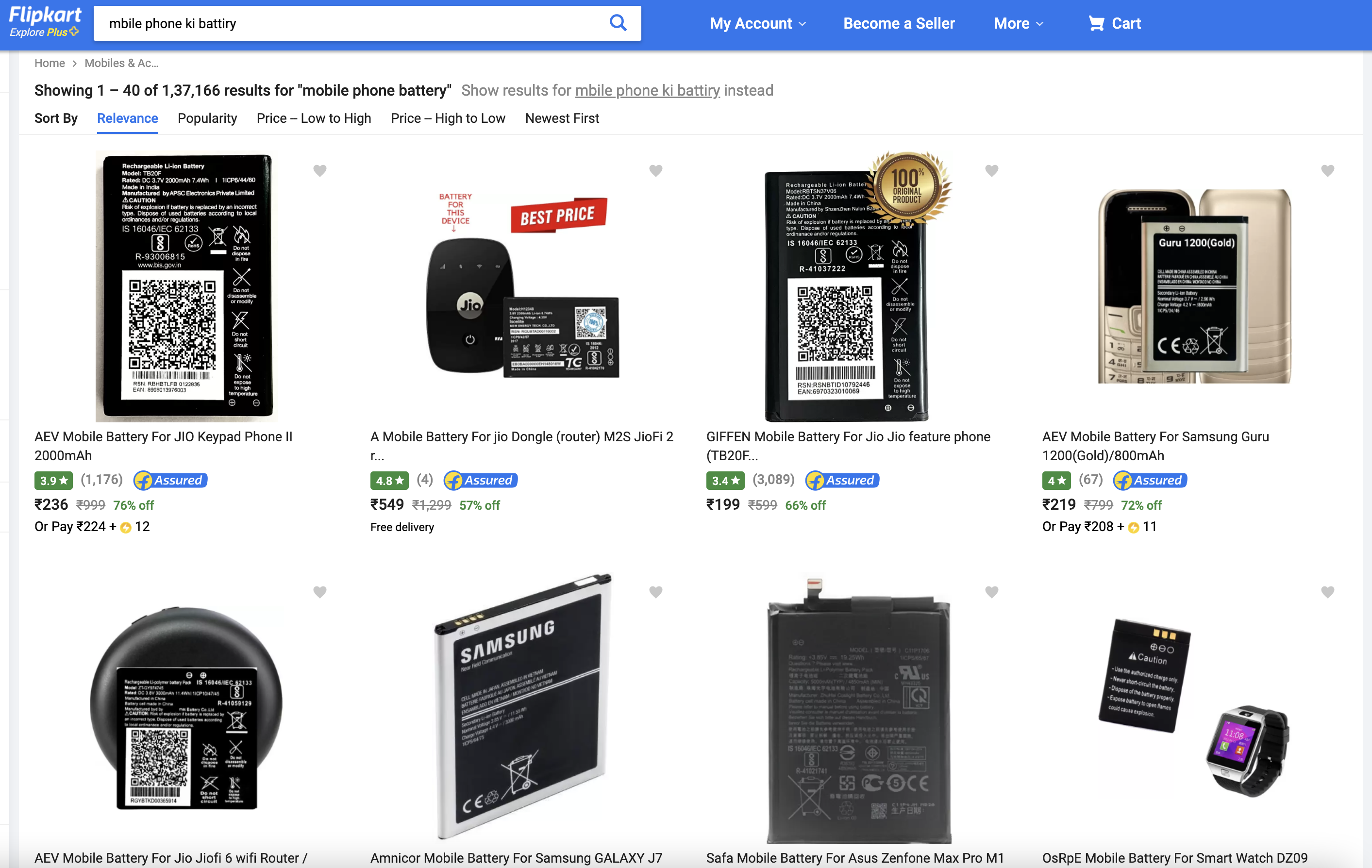}\\\textbf{Query:} kale jute 300 tak &   \textbf{Query:} mbile phone ki battry\\
\includegraphics[width=200pt, height = 150pt]{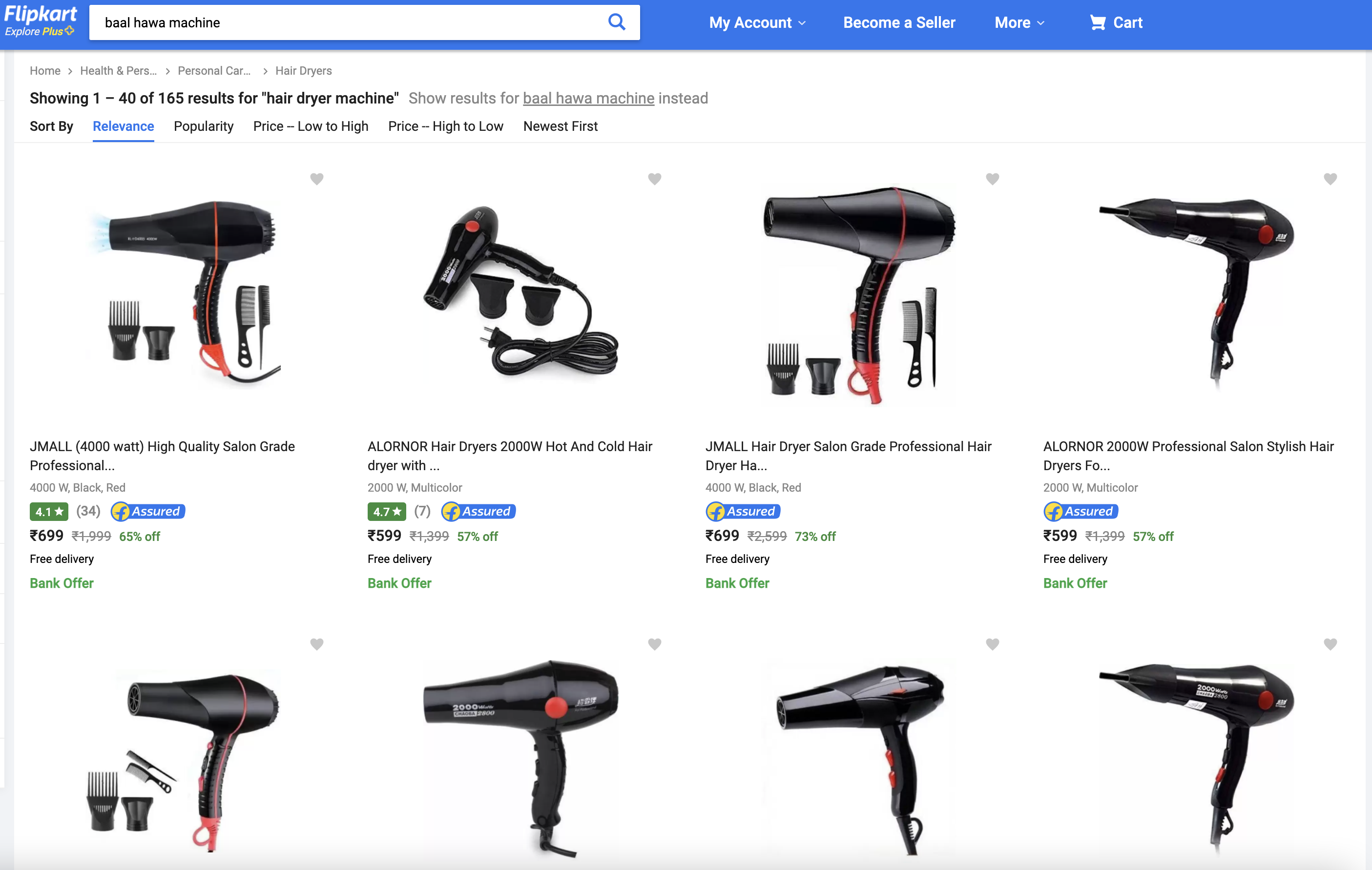}&
\includegraphics[width=200pt, height = 150pt]{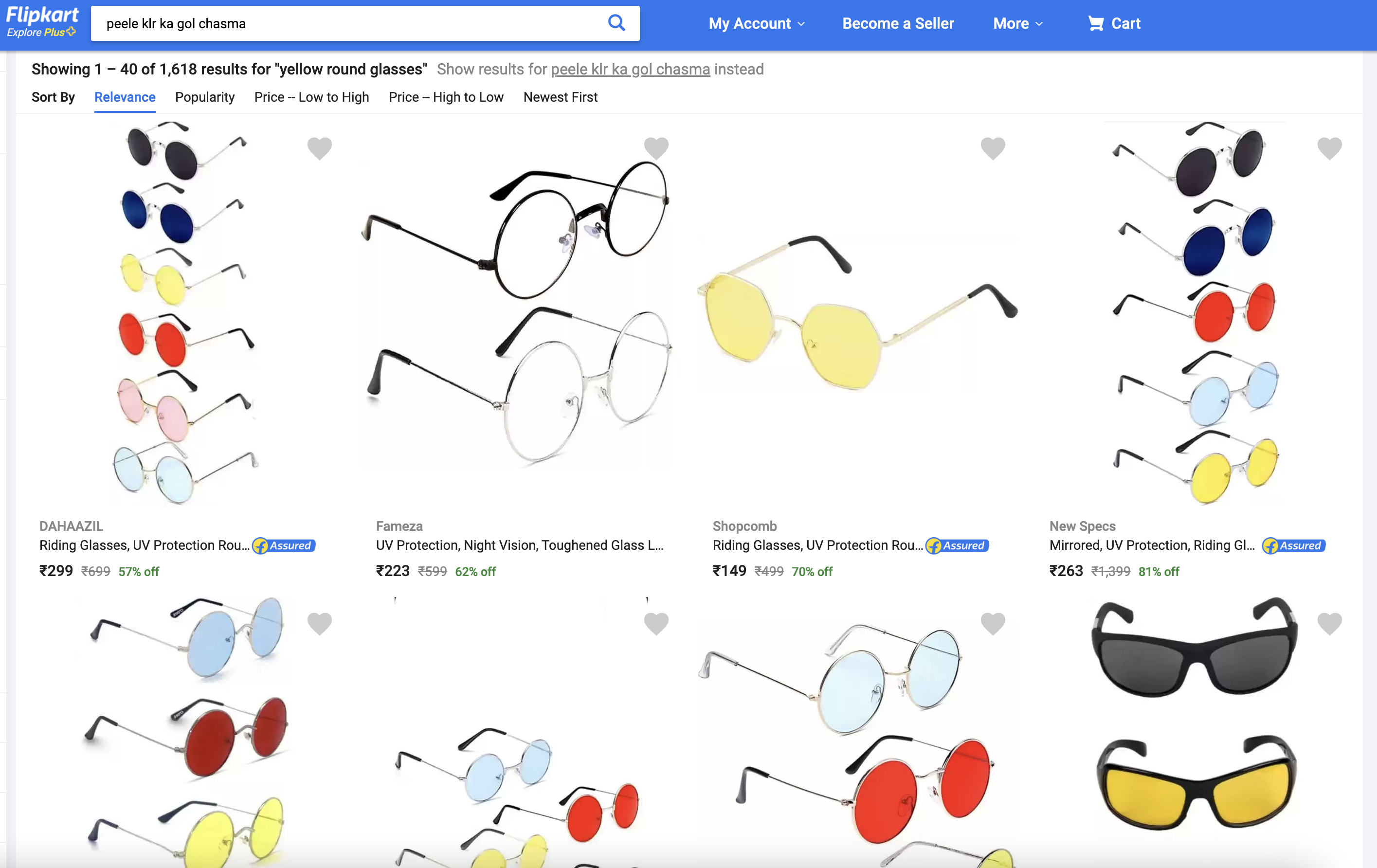}\\\textbf{Query:} baal hawa machine & \textbf{Query:} peele klr ka gol chasma\\

\end{tabular}
\caption{\label{fig:inpt0} Hinglish query search results on Flipkart website.}
\end{figure*}

Table \ref{tab:res2} shows the accuracy result for different KD loss types. JS loss gives a BLEU score improvement of 1.1 over CE and facilitates much better knowledge transfer from teacher to student. Our findings are in line with \cite{englesson2021generalized}.  We calculate latency for all the models on the test set. The best student model provides more than \textbf{4x} speed up for the inference with a ~0.5 drop in BLEU score. We also further optimised the latencies via C libraries for production deployment. % A variant of the student model is currently live in production, enabling search with Hinglish language for millions of queries. 

\section{Production Search Results}

The code-mix translation model is currently live on the Flipkart app and website. Fig. \ref{fig:inpt0} shows the search results for different Hinglish queries. Note that we get good search results even for the queries with spell errors (queries: mbile phone ki battry, peele klr ka chasma), for the queries with articulation gap (query: baal hawa machine), and for the queries where attributes such as color and price are specified (query: kale jute 300 tak).

\section{Additional experiments}

\subsection{Comparison with multilingual pre-trained models}
\label{sec:comp}

We compared the results of BART model with multilingual pre-trained models. Specifically, we consider 2 pre-trained models, mT5-small and mBART-large-50. We follow a similar training procedure for multilingual models where we first fine-tune the model with the 5M corpus for 3 epochs using a learning rate of 5e-6 and then further fine-tune with a manually labeled corpus using a learning rate of 1e-5. Table \ref{tab:res104} shows the comparison accuracy. The result indicates that models pre-trained with English text provide better initialization than multilingual models for query translation. 

\begin{table}[h]
    \centering
    \captionsetup{justification=centering, margin=5mm}
    \begin{tabular}{|c|c|}
        
        \hline
        \textbf{Model} & \textbf{BLEU}\\
        \hline
        BART & 51.3 \\
        \hline
        mT5 & 48.3 \\
        \hline
        mBART & 46.6 \\
        \hline
        
        \end{tabular}
    \vspace{2mm}
    \caption{Multilingual models on Hinglish test set}
    \label{tab:res104}
\end{table}

\subsection{Comparison with in-house generic models for dataset tagging}
\label{sec:inhouse}

We also experimented with building generic translation and transliteration models in-house for dataset tagging and observed comparable results with the generic third-party APIs. Model details below:
\subsubsection{\textbf{Translation model}}

We train a generic translation model for English to Hindi translation. For training the model, we use the following publicly available translation datasets: Opensubtitiles, Pib, PMI, WikiMatrix, ALT, FLORES-101, indian-parallel-corpora, Man ki bat,WikiMedia,Tatoeba, Global voices, GNOME, KDE, Ubuntu. The total size of the training corpus is $\sim$1.4M. 
We train a transformer model with 6 layers in encoder and 6 layers in decoder layers. The model has 12 attention heads, 512 hidden dimension, and $\sim$96M trainable parameters. 
Using this model, we get translations on a large corpus ($\sim$157M) of Hindi monolingual text where text is translated from Hindi to English. The monolingual Hindi text is obtained from datasets such as CC-100, OSCAR, news-crawl, Wikipedia. Using this Hindi-English parallel corpus, we then train an English to Hindi translation model. The model architecture is the same as mentioned earlier in this section. 

\subsubsection{\textbf{Transliteration model}}

For building a transliteration model, we use a dataset of size 96.4k of a word-level manual taggings of Hindi-English transliterations. We train a 2 layer transformer model that uses character level tokenization on this dataset.
The model has 4 attention heads and a hidden dimension size of 128. For transliterating of an input text, a hybrid approach is followed. If a word from input text is present in the manual mappings, the corresponding transliteration is used. If the word is not present, then a transformer model is used with greedy decoding to get the word transliteration. 

To create a large parallel Hinglish-English query corpus, we first translate $\sim$18M unlabeled English search queries to Hindi using the translation model. We then use the transliteration model to transliterate Hindi text to English.

The BART model is then pre-trained on this dataset and finetuned on the manually tagged Hinglish queries as before. Table \ref{table:res124} shows the BLEU score comparison result on the test set. With the dataset created with generic in-house models, we get comparable results as with  third-party APIs. 

% We then used an in-house transliteration model to transliterate Hindi translations to English to get Hinglish-to-English parallel corpus. The in-house transliteration model is built as follows. 

% , we built a generic translation model in-house. We used a transformer model with 6 layers in encoder and 6 layers in decoder layers. The model has approx. 96M trainable parameters. The model is trained with English as the source language and Hindi (Devnagari) as the target language. 

% We created dataset of a mapping of 100k words between Hindi and English. We train a single layer transformer model using the character seq2seq mapping. 

% If the word exist in the mapping, the corresponding is used else, a transformer model's output is used for the word transliteration. 

% 2 layer
% 128 hidden dimension
% 4 attention heads
% character level vocab
% greedy decoding
% 96.4k training data size

% The BART is model is pre-trained on this dataset and finetuned on the manually tagged Hinglish queries. Table \ref{tab:res124} shows the result of the experiment. We get comparable results as with using generic domain third-party APIs for creating the training set.

\begin{table}[!t]
\centering
  \begin{tabular}{ll}
    \toprule
    \multirow{1}{*}{Setting} &  
     \multirow{1}{*}{BLEU} \\
       
      \midrule
      Dataset created with generic domain third-party APIs & 51.3 \\
      Dataset created with in-house generic models & 51.1 \\
    
    \bottomrule
  \end{tabular}
  
  \caption{\label{table:res124} Third-party APIs vs in-house generic models}
  
\end{table}

\subsection{Analysis of Auto-Encoder (AE) objective}
\label{sec:analy}

Language Model (LM) regularization is known to help in the translation tasks \cite{gulcehre2015using}. We hypothesize that Auto-Encoder (AE) augmentation may have a similar effect as adding a LM regularization.
To validate this, we design the following experiment. 
During the training, we analyze the cross attention matrices in different layers in the decoder. This is because the decoder in the conditional generative seq2seq models differs from the decoder in language models (such as GPT) in terms of additional cross attention layers. While generating the output tokens, a conditional decoder also pays attention to encoder states through cross attention layers.
Intuitively,  the conditional decoder with identity cross attention matrices would be equivalent to a language model. To verify if 'any' of the cross attention matrices in the decoder are becoming similar to the identity matrix with AE data augmentation loss, we calculate error using the following equation. 
\begin{eqnarray}\label{eq:so10}
e = \min_{h} || C - I ||_F
\end{eqnarray}
where $C$ indicates the cross attention matrix and $I$ indicates the identity matrix. We calculate the Frobenius norm of the difference between the cross attention matrix and the identity matrix and the error ($e$) is calculated by taking the minimum across the attention heads. Note that, since the same tokenizer is used for the encoder and the decoder, cross attention matrices are square matrices.
The error ($e$) is calculated across all layers in the decoder.
We use the same 200k query dataset as mentioned in section \ref{sec:d} for the experiment. Out of 200k samples, 190k are used for training and 10k are used for validation. The model is trained with the loss as specified in Eq. \ref{eq:so1} where $loss_d$ is only calculated with AE augmentation. The model is trained for 5 epochs and at the end of each epoch, we calculate the average error $e$ on the validation set where target-side data is used as the input and the output as mentioned in section \ref{sec:ae}.

\begin{figure} [!hbt]
\centering
\begin{tabular}{c}

\includegraphics[width=160pt, height = 100pt]{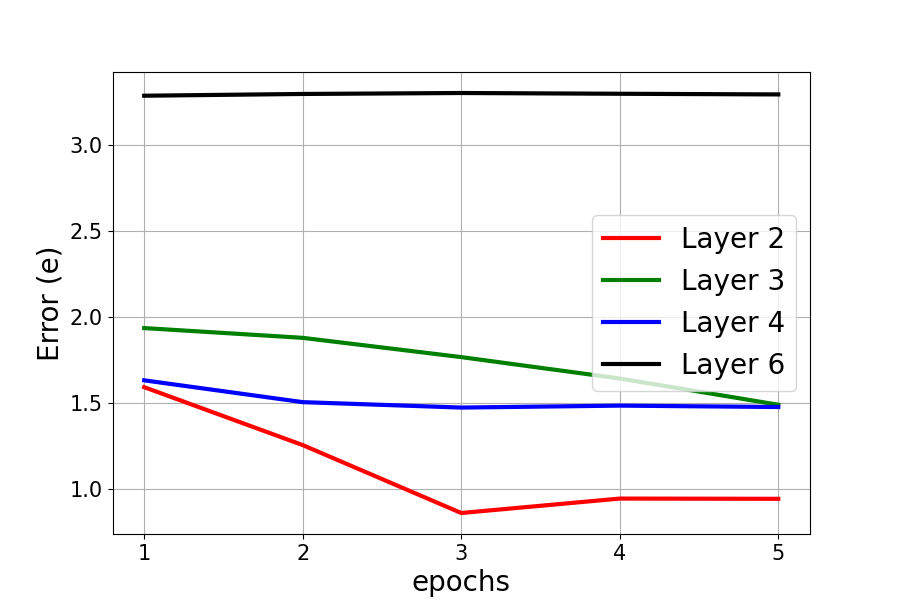}

\end{tabular}
\caption{\label{fig:inpt8} Analysis of cross attention matrices for Auto-Encoder data augmentation.}
\end{figure}

Figure \ref{fig:inpt8} shows the error values for different layers across 5 epochs as the training progresses. Interestingly, for the initial layers (layer: 2,3 4), the error goes on decreasing indicating that at least one of the cross attention matrices is gradually becoming similar to an identity matrix. For the last (6th) layer though, we do not observe a similar trend indicating that the last layer features are more task-specific.

\section{Conclusion}

In the paper, we describe a transfer learning-based approach for code-mix query translation. We found the pre-trained encoder-decoder model and data augmentation technique best suited for the task through experimentation. A larger training corpus is created with translation of the unlabeled query data using generic domain models. To reduce the inference time latency of the model, we used Knowledge Distillation and Weight Quantization. The effectiveness of the proposed method has been validated through experimental evaluations and AB testing. The model is currently live in production, enabling search with Hinglish language for millions of queries.

\section{Acknowledgements}
We would like to thank Amey Patil and Gaurav Gupta for their help in getting results with in-house translation and transliteration models.

\bibliographystyle{ACM-Reference-Format}
\bibliography{sample-base}

\newpage

%\appendix

%\section{Appendix}

\end{document}